\documentclass[runningheads]{llncs}
\usepackage{graphicx}
\usepackage{threeparttable}
\usepackage{orcidlink}
\usepackage{amsmath}
\begin{document}
\title{Knowledge Graphs Quality Assessment in the Era of LLMs: A Conceptual Framework}
\titlerunning{KG Quality Assessment in the Era of LLMs}
%
\author{Elwin Huaman\inst{1}\orcidlink{0000-0002-2410-4977}}

\authorrunning{E. Huaman}

\institute{Centre for Human-Inspired Artificial Intelligence, \\
Institute for Technology and Humanity, University of Cambridge, UK \\
\email{elh97@cam.ac.uk}}

\maketitle              
\begin{abstract}
Knowledge Graphs (KGs) have been popularized during the last few decades and are increasingly constructed and consumed by Large Language Models (LLMs). Various applications therefore rely on KGs to provide concise, complete, accurate, faithful, fresh, and traceable answers. However, what is the quality of those KGs? In which cases should a KG be used, and can it be trusted to ground an LLM? How might it be evaluated? We reviewed the literature on quality assessment (QA) of data, information, linked data, KGs, and the intersection of KGs and LLMs. We extended the current state-of-the-art frameworks to a catalog of 24 quality dimensions (QDs) and various quality metrics (QMs). Furthermore, we propose a general-purpose, customizable to a domain or task, and practical QA framework for assessing the quality of KGs.
\keywords{Knowledge Graph Quality \and Assessment \and LLMs}
\end{abstract}
\section{Introduction}
\label{sec:introduction}
Knowledge Graphs (KGs) have gained popularity over the last few decades. In 2012, the Google's Knowledge Graph was presented, and in the same year, Wikidata KG was released as crowd-sourced, non-commercial, and open on the web. Furthermore, many other KGs are available on the web, either created semi-automatically (e.g. DBpedia) or automatically (e.g., NELL). However, an issue that comes with KGs is their quality, a context that has changed substantially with the advent of Large Language Models (LLMs). On the one hand, LLMs are increasingly used to create, populate, and curate KGs, e.g., through ontology learning \cite{BabaeiDA2023}, triple extraction \cite{ZhangS24}, and automated validation of statements \cite{AdamK25,BoylanSHTGM24}, introducing quality risks that were less common in human-curated settings: a statement can appear well-formed yet hallucinated (unfaithful) \cite{BoylanSHTGM24,ShamiMS26}.
On the other hand, KGs are increasingly consumed by LLMs as ground truth in retrieval-augmented generation, where they measurably reduce hallucinations and improve answer accuracy \cite{AliTM26,SequedaAJ24,WangQZYPBLDQ25}.

Quality involves several dimensions, such as accessibility, accuracy, appropriate amount, and so on. In terms of KGs and LLMs, few works have been proposed. To address this gap, we have looked at the literature on quality assessment (QA) and identified 24 quality dimensions (QDs) and several quality metrics (QMs) based on the Goal Question Metric (GQM) approach. 

We propose a general-purpose, customizable to a use case or task, and practical QA framework for KGs. Our approach involves (1) identifying users, use case, and KGs, (2) selection of QDs and QMs by means of weights of importance, (3) performing the assessment of QMs, QDs, and total score for a KG, and (4) exploitation of the results.

%
\section{Quality Assessment (QA)}
\label{sec:quality-assessment}
The most relevant work on QA over the past years involves quality of data, information, linked data, KGs, and KG use cases. In order to accurately define and measure a QD, it is not enough to review the quality of data or linked data solely. In fact, KG quality needs an overall picture of QA within the years of their popularization. Defining what quality is, within the context of KGs and LLMs, will depend greatly on whether QDs have been identified for the data consumers~\cite{WangS96}, the data source~\cite{bizer2007,NaumannR00}, and the use case~\cite{FarberBMR18,FenselSAHKPTUW20,WangQZYPBLDQ25,ZaveriRMPLA16}.

\begin{table}
\caption{Comparison of QDs proposed by authors.}
\label{tab:quality-dimensions}
\tiny
\centering
\begin{tabular}{|p{1cm}|p{3.1cm}|p{7.7cm}|}
\hline
Author & Summary & Dimensions \\ 
\hline
Wang and Strong, 1996~\cite{WangS96} & 15 QDs grouped in 4 categories: Accessibility (2), Contextual (5), Intrinsic (4), Representational (4) & Accessibility, Accuracy, Appropriate amount, Believability, Completeness, Concise representation, Consistent representation, Ease of understanding, Interpretability, Objectivity, Relevancy, Reputation, Security, Timeliness, Value added \\ 
\hline
Naumann and Rolker, 2000~\cite{NaumannR00} & 22 QDs grouped in 3 categories: Object criteria (9), Process criteria (6), Subject criteria (7) & Accuracy, Amount of data, Availability, Believability, Completeness, Concise representation, Consistent representation, Customer support, Documentation, Interpretability, Latency, Objectivity, Price, Relevancy, Reliability, Reputation, Response time, Security, Timeliness, Understandability, Value added, Verifiability \\ 
\hline
Bizer, 2007~\cite{bizer2007} & 16 QDs grouped in 4 categories: Accessibility (2), Contextual (7), Intrinsic (4), Representational (3) & Accessibility, Accuracy, Appropriate amount, Believability, Completeness, Concise representation, Consistency, Consistent representation, Interpretability, Objectivity, Offensiveness, Relevancy, Response time, Timeliness, Understandability, Verifiability \\ 
\hline
Zaveri et al., 2016 \cite{ZaveriRMPLA16} & 18 QDs grouped in 4 categories: Accessibility (5), Contextual (4), Intrinsic (5), Representational (4) & Accuracy, Availability, Completeness, Concise representation, Conciseness, Consistency, Interlinking, Interoperability, Interpretability, Licensing, Performance, Relevancy, Security, Syntactic validity, Timeliness, Trustworthiness, Understandability, Versatility \\ 
\hline
F\"{a}rber et al., 2018 \cite{FarberBMR18} & 11 QDs in 4 categories: Accessibility (3), Contextual (3), Intrinsic (3), Representational (2) & Accessibility, Accuracy, Completeness, Consistency, Interlinking, Interoperability, Licensing, Relevancy, Timeliness, Trustworthiness, Understandability \\ 
\hline
Fensel et al., 2020 \cite{FenselSAHKPTUW20} & 23 QDs & Accessibility, Accuracy, Appropriate amount, Believability, Completeness, Concise representation, Consistent representation, Cost effectiveness, Ease of manipulation, Ease of operation, Ease of understanding, Flexibility, Free of error, Interoperability, Objectivity, Relevancy, Reputation, Security, Timeliness, Traceability, Understandability, Value added, Variety \\
\hline
Hogan et al., 2021 \cite{Hogan21} & 10 QDs in 4 categories: Accuracy (3), Coverage (2), Coherency (2), Succinctness (3) & Accuracy, Coverage, Coherency, Succinctness, Interoperability, Licensing, Relevancy, Timeliness, Trustworthiness, Understandability \\ 
\hline
Xue et al., 2023 \cite{XueZ2023} & 17 QDs in 4 categories: Accessibility (4), Contextual (4), Intrinsic (5), Representation (5) & Accuracy, Availability, Completeness, Conciseness, Consistency, Diversity, Interlinking, Interpretability, Licensing, Quantity, Redundancy, Relevancy, Security, Timeliness, Trustworthiness, Understandability, Versatility \\ 
\hline
Wang et al., 2025 \cite{WangQZYPBLDQ25} & 23 QDs in 5-stage categories: Input source (4), Construction process (3), Achieve content (8), Application condition (2), Access path (6) & Accuracy, Availability, Completeness, Conciseness, Confidence, Interlinking, Interoperability, Interpretability, Licensing, Logical Consistency, Perspective Diversity, Recall, Reliability, Representativeness, Scale Diversity, Security, Semantic Accuracy, Spatial, Temporal Accuracy, Time Validity, Topology Consistency, Transmission, Uniqueness \\ 
\hline
\end{tabular}
\end{table}
Table~\ref{tab:quality-dimensions} shows how QDs evolved across data~\cite{WangS96}, information~\cite{bizer2007,NaumannR00}, linked data~\cite{FarberBMR18,ZaveriRMPLA16}, and KGs~\cite{FenselSAHKPTUW20,Hogan21,XueZ2023,WangQZYPBLDQ25}. 
The most cited work from this group, Wang and Strong~\cite{WangS96}, empirically collected and organized QDs from data consumer perspectives. They point out the importance of measuring quality to the extent to which data is accessible, contextually dependent on the task at hand, clearly represented, and intrinsically largely correct and complete. Naumann and Rolker~\cite{NaumannR00} group QDs into 3 categories (subject, predicate, object). The subject (or subject-criteria) implies the user perspective to evaluate specific QDs, the predicate (or process-criteria) implies mainly the process of querying (e.g., availability), and the object (or object-criteria) involves the knowledge source itself (e.g., completeness). Compared with Wang and Strong, Naumann and Rolker highlight the importance of availability, price, customer support, documentation, latency, reliability, response time, understandability, and verifiability QDs. 

Bizer~\cite{bizer2007} argues that quality is a multidimensional concept that involves 3 main aspects: the data, its context, and its trustworthiness. The author proposes consistency and offensiveness as additional QDs compared with ~\cite{NaumannR00,WangS96}. Furthermore, Bizer contemplates offensiveness as a subjective QD that needs to be taken into account in specific contexts (e.g., cultural).
Zaveri et al.~\cite{ZaveriRMPLA16} discuss QA for linked data extensively. They identified 18 QDs and several assessment tools. Moreover, compared with \cite{bizer2007,NaumannR00,WangS96}, Zaveri et al., introduce additional QDs such as conciseness, interlinking, interoperability, licensing, performance, syntactic validity, trustworthiness, and versatility.
F\"{a}rber et al.~\cite{FarberBMR18} study the state of the art on QA and implement 11 QDs, which are evaluated through 34 QMs. Moreover, the authors test their framework against publicly available KGs (e.g., DBpedia, Wikidata), analyse their results, and provide recommendations on when to use each KG.
Fensel et al.~\cite{FenselSAHKPTUW20} provide a list of 23 QDs to assess KGs. Compared with~\cite{bizer2007,FarberBMR18,NaumannR00,WangS96,ZaveriRMPLA16}, the authors propose cost-effectiveness, ease of manipulation, ease of operation, flexibility, free of error, traceability, and variety QDs. Furthermore, Hogan et al.~\cite{Hogan21} provide a list of 10 QDs described in \cite{ZaveriRMPLA16}.

More recently, Xue and Zou \cite{XueZ2023} survey KG quality management comprehensively, covering 17 QDs, QMs, and the management processes of assessment like error detection, correction, and completion.
Wang et al., \cite{WangQZYPBLDQ25} propose a Quality Assessment Framework for geographic KGs with lifecycle-organized QMs and a streamlined weighted schema, explicitly showcasing the needs of training and grounding geographic LLMs. In parallel, LLM-specific QA literature has emerged that proposes metrics such as faithfulness, answer relevance, and context relevance for retrieval-augmented generation \cite{ESJAS24}, benchmarks for validating KG statements with LLMs \cite{AdamK25,BoylanSHTGM24}. A goal of this paper is to connect these two strands: LLMs' and KGs' quality assessment.

Most of the authors, except \cite{WangS96,ZaveriRMPLA16}, do not describe the methodology behind their QDs, yet they were created in the context the authors focus on. We build on the QDs listed in Table~\ref{tab:quality-dimensions}: Flexibility relates to the ease of manipulation dimension~\cite{FenselSAHKPTUW20}, understandability is also called ease of understanding~\cite{ZaveriRMPLA16}, and value-added is also considered as part of completeness~\cite{FarberBMR18}. Since KGs are characterized by their variety across domains~\cite{FarberBMR18,FenselSAHKPTUW20,Paulheim17} and since cost~\cite{FenselSAHKPTUW20,Lenat95,HuamanF21,WangS96}, traceability~\cite{FenselSAHKPTUW20,HuamanKF2020,HuamanTF21}, faithfulness \cite{BoylanSHTGM24,ShamiMS26}, and contextual validity \cite{WangGY2026} QDs have gained attention in the LLMs context, we summarize our findings into 24 QDs relevant to KGs and LLMs, see Table~\ref{tab:quality-goals-questions-metrics}.

\section{Assessing Knowledge Graph Quality}
\label{sec:assessing-kg-quality}
To operationalize the measurement of the 24 QDs, we employ the Goal Question Metric (GQM) approach, which defines (i) a goal, (ii) questions to achieve it, and (iii) metrics to answer them. Originally developed for software quality, GQM has also been applied to data quality~\cite{Redman0093293} and Linked Open Data~\cite{BehkamalKBJ14}. We classify each metric as quantitative (QN), where a value score can be calculated manually, semi-automatically, or automatically, or qualitative (QL), where user judgment is required.
Table~\ref{tab:quality-goals-questions-metrics} displays the 24 QDs with their questions, metrics, and classification. These definitions might not satisfy every use case or domain and can be adapted accordingly.

\begin{table}
\caption{Proposed Goals and Questions for assessing the quality of KGs.}
\label{tab:quality-goals-questions-metrics}
\tiny
\centering
\begin{threeparttable}
\begin{tabular}{|p{2.5cm}|p{9.2cm}|}
\hline
Goal & Question, Metrics, and Measurement Type (Quantitative:QN and Qualitative:QL) \\ 
\hline
Accessibility & Is the KG (or at least part of it) available (QN), provides a SPARQL endpoint (QN), retrievable (QN), supports content negotiation (QN), and contains a license (QN)? \\
\hline
Accuracy & Is the KG reliable and correct, e.g., syntactically (QN) and semantically (QN)? Are LLM-generated triples verified against external evidence (QN)? \\ 
\hline
Appropriate amount & Does the KG contain an appropriate amount (QN) of instances for a specific task and context? \\ 
\hline
Believability &  Does the KG provide provenance information (QN), is trustworthy (QL), and has no unknown or empty values (QN)? Do LLM-generated triples carry their supporting evidence and a trust rating (QN)? \\ 
\hline
Completeness & To what degree is the KG complete regarding data (QN), population (QN), and interlinking (QN)? To what degree does the KG cover long-tail entities (QN)? \\ 
\hline
Concise representation & Is the KG concisely represented by avoiding blank nodes (QN) and reification (QN)?\\ 
\hline
Consistent representation & Is the KG consistently represented, e.g., disjoint inconsistencies of classes (QN), inconsistent inverse functional property values (QN), schema restrictions (QN)? \\ 
\hline
Contextual validity & Does the KG represent the conditions under which a statement holds, e.g., temporal (QN), spatial (QN), and situational (QL) validity?\\
\hline
Cost-effectiveness &  Does the KG require extra data at any cost (QL), e.g., inference cost (QN)? \\ 
\hline
Ease of manipulation & At which level does the KG provide documentation (QL)?   \\ 
\hline
Ease of operation & Is it possible to update (QN), download (QN), and integrate (QN) the KG?  \\ 
\hline
Ease of understanding & Is the KG represented using self-descriptive URIs (QN) and in various languages (QN)? Can triples be verbalised as model prompts (QN)? \\ 
\hline
Faithfulness & Are the KG's statements supported by their attached sources (QN), and are outputs generated from the KG grounded in the retrieved statements (QN)? \\ 
\hline
Free of error & Does the KG provide correct property values (QN) and is it consistent with its source text (QN)? \\ 
\hline
Grounding effectiveness & To what degree does the KG improve the reliability of a consuming model, e.g, answer-accuracy uplift (QN) and hallucination-rate reduction (QN)? \\
\hline
Interoperability &  Is the KG interoperable, e.g., openly available (QN) and uses standard vocabularies (QL)?  \\ 
\hline
Objectivity &  Is the KG objective, e.g., unbiased (QL) and declares provenance information (QN)? Has bias been audited (QL)?\\ 
\hline
Relevancy & Is the KG relevant for the task at hand, e.g., at which level the KG provides knowledge for a specific domain or use case (QL)? For KG-grounded applications: context relevance of retrieved subgraphs per query (QN)? \\ 
\hline
Reputation & Is the KG well rated? E.g., is the KG well positioned in explicit ratings (QL)? \\ 
\hline
Security &  Does the KG provide security mechanisms like digital signature (QN) and KG authentication (QL)?  \\
\hline
Structural fidelity & Does the KG preserve the conceptual organization of its domain, e.g., structural similarity to a reference ontology (QN)? Classes, properties, and schemas (QN)? To what degree the KG has been human-curated, LLM-generated, or hybrid (QN)? \\
\hline
Timeliness & To what degree is the KG up to date (QN) and fresh (QN)? How does its quality evolve under change (QL)?\\ 
\hline
Traceability & Does the KG provide means to verify its provenance (QL) at statement level (QN) and authenticity (QL)? Is the provenance automatically verifiable (QN)?\\ 
\hline
Variety & To what degree does the KG integrate various domain sources (QL)? \\
\hline
\end{tabular}
\end{threeparttable}
\end{table}

\section{Knowledge Graph Quality Assessment Framework}
\label{sec:kg-quality-framework}
\begin{figure}[h]
\includegraphics[width=\textwidth]{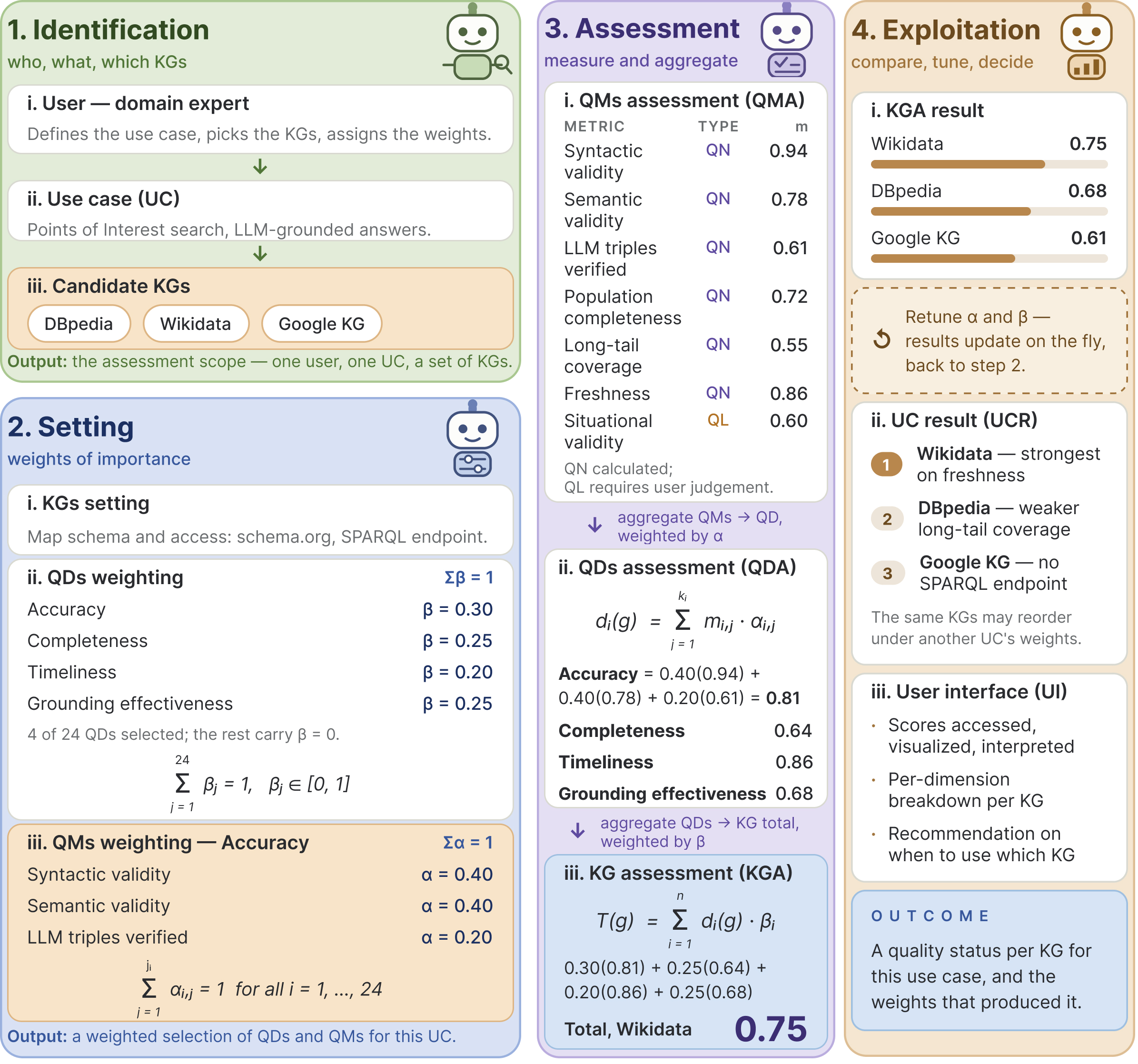}
\caption{The KG Quality Assessment Framework on an illustrative use case (a Points-of-Interest): A domain expert (1) fixes the UC, and candidate KGs; (2) sets KGs and weights for QDs and QMs; A system (3) assesses and aggregates them into $d_i(g)$ and $T(g)$; The user (4) compares, tunes, and decides about the quality of candidate KGs.} 
\label{fig:kgqa-framework}
\end{figure}

We propose a user-driven assessment framework in which users select QDs and QMs according to their importance for a domain, task, or use case. It consists of 4 steps: identification, setting, assessment, and exploitation (see Fig.~\ref{fig:kgqa-framework}).

\subsection{Identification}
\label{subsec:identification}
In this step, we propose to identify 3 main points: a user, a use case, and a KG.
\begin{enumerate}
    \item[i.] \textit{Users} (domain experts) guide the definition of the use case, the KGs to evaluate, and the weights for QDs and QMs. In the LLMs context, a user may also be an LLM-based application (``consumer'').
    \item[ii.] \textit{Use Cases} (UCs) determine how weights are assigned. A KG can be a product of, or serve, an LLM application, and may be assessed under both roles with different weight profiles.
    \item[iii.] \textit{KGs} relevant to the UC are selected, e.g., DBpedia, Google's Knowledge Graph, and Wikidata for the Points of Interest (POI) domain, see Fig.~\ref{fig:kgqa-framework}.
\end{enumerate}
\subsection{Setting}
The goal of developing an assessment framework is to obtain a quality status of a KG for a specific UC. For that, the user sets up each KG in the framework and selects QDs and QMs by weights of importance. This process involves:
\begin{enumerate}
    \item[i.] \textit{KGs Setting}. KGs use heterogeneous schemas (e.g., \texttt{so:name}\footnote{\texttt{so:name} (\url{https://schema.org/name})}, \texttt{wd:Q82799}\footnote{\texttt{wd:Q82799} (\url{https://www.wikidata.org/wiki/Q82799})}) and access methods (e.g., SPARQL, API) so they must be adapted per KG.
    \item[ii.] \textit{QDs Weighting}. Beta-weights (\(  \beta_{i} \)) are defined for selecting QDs, such that \(  \beta_{i} \) defines a weight of importance for each QD and \(  \beta_{i} \in  \left[ 0,1 \right]  \), where \( 0 \) is the minimum degree of importance and a value \( 1 \) is the maximum degree.
    \item[iii.] \textit{QMs Weighting}. Alpha-weights (\(\alpha_{i,j}\)) are defined for selecting QMs, such that \(\alpha_{i,j} \in [0,1]\) expresses the degree of importance of the $j^{th}$ QM \(m_{i,j}\) of the $i^{th}$ QD. Furthermore, both weight vectors must be normalized:
    \[
      \sum_{i=1}^{24}\beta_i = 1
      \qquad\text{and}\qquad
      \sum_{j=1}^{k_i}\alpha_{i,j} = 1 \;\; \text{for all } i=1,\dots,24,
    \]
    where \(k_i\) is the number of QMs selected for the \(i\)th QD.
\end{enumerate}
\subsection{Assessment}
In this step, the framework performs 3 main tasks: assessing the QMs, calculating the aggregated score per QD, and computing a total aggregated score per KG. 
\begin{enumerate}
    \item[i.] \textit{QMs Assessment} (QMA) executes each QM manually, semi-automatically, or automatically, tagged as QN (computed) or QL (human judgment) measure.
    \item[ii.] \textit{QDs Assessment} (QDA) is represented by $d_{i}(g)$, which aggregates the QMA values of the $i^{th}$ QD for a KG $g$, where $d_{i}(g) \in [0,1]$.
    \item[iii.] \textit{KG Assessment (KGA)} is represented by $T(g)$, which calculates the total aggregated score from the values of each QDA for a KG $g$. 
     \[ 
        d_{i}(g) = \sum_{j=1}^{k_{i}}{m_{i,j}}\cdot\alpha_{i,j} 
        \qquad\text{and}\qquad
        T(g) = \sum_{i=1}^{24}{d_{i}(g)}\cdot\beta_{i},
     \] 
    where $m_{i,j}$ is the score of the $j^{th}$ QM of the $i^{th}$ QD,
    and $m_{i,j}$, $T(g) \in [0,1]$.
\end{enumerate}

\subsection{Exploitation}
As described above, the alpha- and beta- weights affect the overall result of an assessed KG. Therefore, the results can be tuned, compared, and visualized.
\begin{enumerate}
    \item[i.] \textit{KGA result}. Users may tune weights and recompute scores on the fly.
    \item[ii.] \textit{UC Result} (UCR). Users compare and rank KGA results per UC.
    \item[iii.] \textit{User Interface} (UI). It provides means to access, visualize, and interpret the results. For instance, giving recommendations on when to use a KG. 
\end{enumerate}
The framework's proposed QDs and QMs reflect our interpretation and are meant to be adapted per UC. Since some QMs overlap across QDs, users should select a relevant subset and tune the weights until the profile reflects their UC.
\section{Conclusion and Future Work}
\label{sec:conclusion}
We conducted a study on quality assessment, which included data, information, linked data, KGs, and LLMs, proposing 24 QDs and several QMs, highlighting cost-effectiveness, faithfulness, and contextual validity. Furthermore, we propose a user-driven, four-step assessment framework: i) identification, ii) setting, iii) assessment, and iv) exploitation, which can guide KG architects on quality requirements and inform future assessment frameworks. 
As future work, the framework can be semi-automated by implementing the quantitatively measurable (QN) metrics, reducing assessment costs while retaining human-machine collaboration, and should be evaluated in specific use cases. For instance, it might support the correct adoption of KGs within LLM workflows.
\bibliographystyle{splncs04}
\bibliography{bibliography}

\end{document}